\documentclass[a4paper,twoside]{article}
\usepackage{graphicx}%
\usepackage{multirow}%
\usepackage{amsmath,amssymb,amsfonts}%
\usepackage{mathrsfs}%
\usepackage{xcolor}%
\usepackage{textcomp}%
\usepackage{manyfoot}%
 \usepackage{listings,url}%
\usepackage{epsfig}
\usepackage{subcaption}
\usepackage{calc}
\usepackage{amssymb}
\usepackage{amstext}
\usepackage{amsmath}
\usepackage{amsthm}
\usepackage{multicol}
\usepackage{pslatex}
\usepackage{apalike}
\usepackage{algorithm2e}
\usepackage{SCITEPRESS} 

\newcommand{\eq}[1]{\begin{align}#1\end{align}}
\newcommand{\eqs}[1]{\begin{align}#1\end{align}}

\begin{document}

\title{A Survey of Deep Learning: From Activations to Transformers}

\author{\authorname{Johannes Schneider\sup{1} and Michalis Vlachos\sup{2}}
\affiliation{\sup{1} University of Liechtenstein, Vaduz, Liechtenstein}
\affiliation{\sup{2} University of Lausanne, Lausanne, Switzerland}
\email{johannes.schneider@uni.li, michalis.vlachos@unil.ch}
}




\abstract{Deep learning has made tremendous progress in the last decade. A key success factor is the large amount of architectures, layers, objectives, and optimization techniques. They include a myriad of variants related to attention, normalization, skip connections, transformers and self-supervised learning schemes -- to name a few. We provide a comprehensive overview of the most important, recent works in these areas to those who already have a basic understanding of deep learning. We hope that a holistic and unified treatment of influential, recent works helps researchers to form new connections between diverse areas of deep learning. We identify and discuss multiple patterns that summarize the key strategies for many of the successful innovations over the last decade as well as works that can be seen as rising stars. We also include a discussion on recent commercially built, closed-source models such as OpenAI's GPT-4 and Google's PaLM 2.}


\onecolumn \maketitle \normalsize \setcounter{footnote}{0} \vfill

\section{Introduction} 
Deep learning is widely regarded as the driving force behind artificial intelligence. Its models have achieved top leaderboard rankings in various fields, including computer vision, speech, and natural language processing. One of the major advantages of deep learning is its layered, modular structure, which allows for the construction of models from individual components in a flexible manner. Researchers have created a large selection of layers, architectures, and objectives. Keeping up with the ongoing developments in the various aspects of deep learning is a difficult task. Although specific surveys are available, there is currently no comprehensive overview of recent progress covering multiple aspects of deep learning such as learning, layers and architecture. There exist multiple reviews with a narrow focus such as large language models ( e.g. \cite{min21}) and convolutional neural networks (e.g. \cite{kha20}). Previous studies \cite{alom19,shre19,dong21,alz21} with a wider focus have often overlooked new developments such as transformers and supervised-learning. However, taking a more comprehensive and more holistic look at various disciplines can be extremely advantageous: For example, NLP and computer vision have often influenced each other; CNNs were initially introduced in computer vision, but were later applied in NLP, while transformers were introduced in NLP and later adapted in computer vision. Therefore, removing barriers between disciplines can be highly beneficial. This paper takes this motivation by surveying the recent progress of deep learning from a holistic standpoint, rather than focusing on a particular niche area. We also believe that this is a necessary step, since major innovations have slowed down in terms, i.e., now most architectures are based on the transformer architecture, which dates back to 2017\cite{Scale17}.

It is difficult, if not impossible, to provide an encompassing overview of the field due to the sheer number of articles published yearly and the continual increase in relevant topics, such as transformers and self-supervised learning that have become popular only recently. Our strategy is to choose influential works through (i) usage statistics and (ii) specialized surveys. We also offer an invigorating discussion of shared design patterns across areas that have been successful.




\begin{figure*}[h]
  \centering
\includegraphics[width=0.5\linewidth]{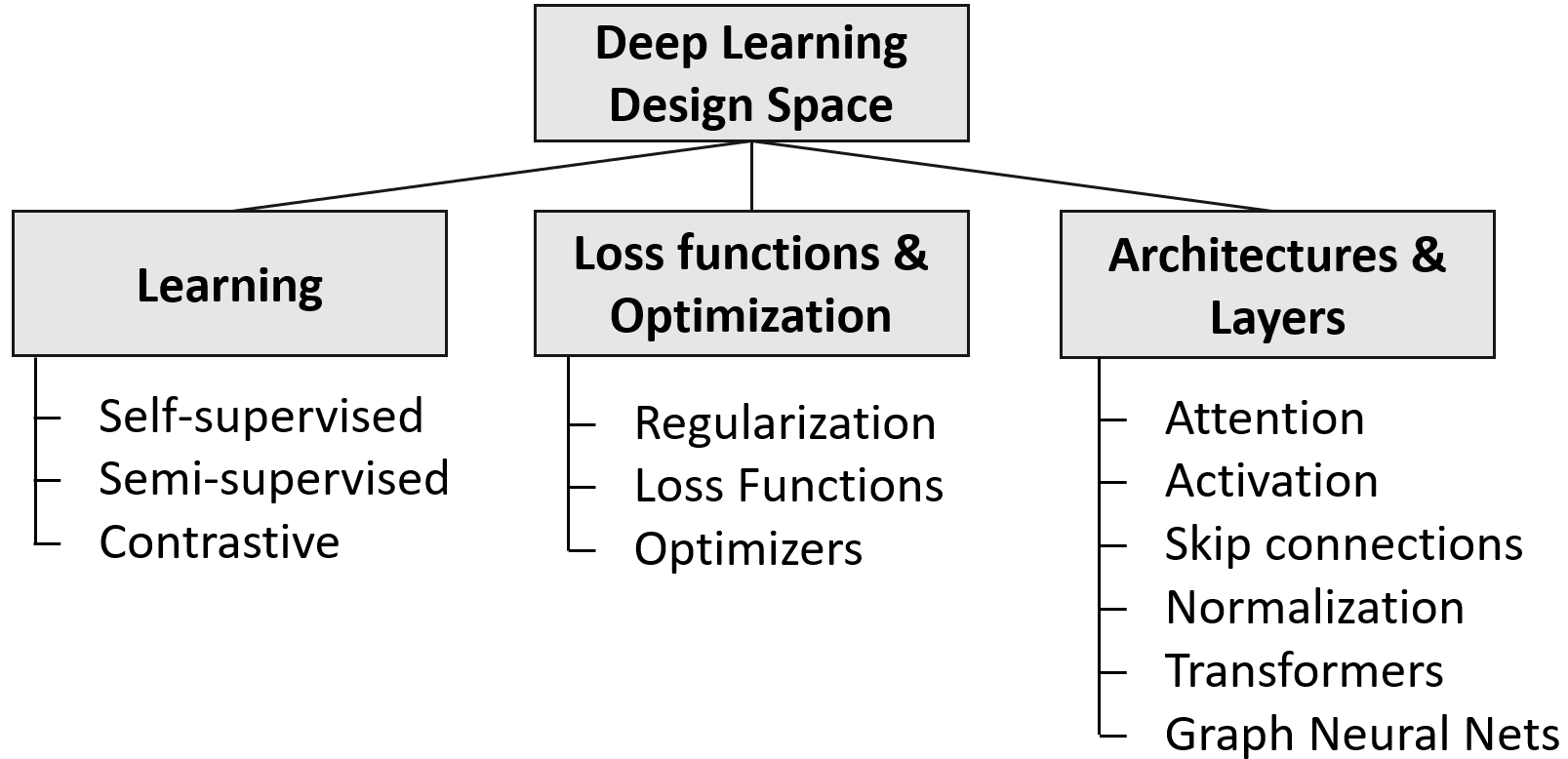}
\caption{Categorization of deep learning and areas covered in the survey} \label{fig:over}
\end{figure*}

\smallskip

\section{Overview} 
Figure \ref{fig:over} provides an overview of the areas included in this survey. We have investigated deep learning design, including objectives and training. We have also given special attention to works that have been somewhat established based on the usage statistics from the popular platform "Paperswithcode.com." There has been an increase in these types of platforms that enable the upload of papers (and models) and provide information on citations, as well as leaderboards. Although there are drawbacks when utilizing data from these platforms, we believe that it offers a new perspective compared to traditional survey methods that often select more arbitrarily. We have only included a selection of the most influential works published from 2016 onwards, as well as rising stars (from 2020 or newer) that have gained significant popularity in a short time.

The extent to which each topic is covered depends on the amount of recent research that has been conducted and its foundational nature. We do not discuss data or computational aspects such as data augmentation, model compression, and distributed machine learning. As a result of limited space, we had to be selective when it came to model families and left out relevant ones such as multi-modal models and autoencoders. 



\smallskip

\section{Loss functions and Optimization}
We discuss common loss functions and optimizers.

\smallskip

\subsection{Loss Functions}
Loss functions (surveyed in \cite{wan20}) often consist of multiple terms that are enhanced with a regularization term. Loss functions are often task-specific but some general ideas are applicable across tasks. Commonly, multiple loss terms are aggregated in a weighted manner. Many papers improve prior work (simply) by using a different loss function. 
 
\smallskip
\noindent The \textbf{Triplet Loss} \label{sec:Tripl}
\cite{Tripl18} was introduced for Siamese networks (Its origin dates back further \cite{schu03}.) The high level idea is to compare a given input to a positive and a negative input and maximize association between positively associated inputs, while minimizing those of negative ones.  It takes input pairs $(x,y)$, each processed by a separate but identical network. It maximizes the joint probability $p(x,y)$ of all pairs $(x,y)$:
\eq{L(\mathcal{V}_{p},\mathcal{V}_{n}) &=-\frac{1}{|\mathcal{V}_{p}|\cdot |\mathcal{V}_{n}|} \sum_{x \in \mathcal{V}_{p}}\sum_{y \in \mathcal{V}_{n}} \log p(x,y)  \\
&=-\frac{1}{|\mathcal{V}_{p}|\cdot |\mathcal{V}_{n}|} \sum_{x \in \mathcal{V}_{p}}\sum_{y \in \mathcal{V}_{n}} \log (1+e^{x-y}) } \\
\noindent Here, $\mathcal{V}_{p}$ and $\mathcal{V}_{n}$ are the positive and negative score set respectively.

\smallskip
\noindent\textbf{Focal Loss}\cite{Focal17} \label{sec:Focal}
focuses learning on hard misclassified samples by altering the cross entropy loss. It adds a factor $(1-p)^{\gamma}$, where $p$ denotes the probability of a sample stemming from the cross entropy loss and $\gamma$ is a tunable parameter.
\eq{L(p)=(1-p)^{\gamma}\log(p)}

\smallskip
\noindent The \textbf{Cycle Consistency Loss}\cite{Cycle17} \label{sec:Cycle}
 is tailored towards unpaired image-to-image translation of generative adversarial networks. For two image domains $X$ and $Y$, the loss supports the learning of mappings $G: X \rightarrow Y$ and $F: Y \rightarrow X$ so that one reverses the other, i.e., $F(G(x))\approx x$ and $G(F(y)) \approx y$.
\eqs{L(G,F)&= \mathbb{E}_{x \sim p_{data}(x)}[||F(G(x)) - x||_{1}]\\
&+ \mathbb{E}_{y \sim p_{data}(y)}[||G(F(y)) - y||_{1}]}




\smallskip
\noindent The \textbf{Supervised Contrastive Loss}\cite{Super20}  \label{sec:Super}
pulls together clusters of points of the same class in embedding space and pushes samples of different classes apart. It aims at leveraging label information more effectively than cross-entropy loss.
\eqs{\mathcal{L}_i^{sup}&=\frac{-1}{2N_{\boldsymbol{\tilde{y}}_i}-1}\cdot \\
&\sum_{j=1}^{2N}\mathbf{1}_{i\neq j}\cdot\mathbf{1}_{\boldsymbol{\tilde{y}}_i=\boldsymbol{\tilde{y}}_j}\cdot\log{\frac{\exp{(\boldsymbol{z}_i\cdot\boldsymbol{z}_j/\tau)}}{\sum_{k=1}^{2N}\mathbf{1}_{i\neq k}\cdot\exp{(\boldsymbol{z}_i\cdot\boldsymbol{z}_k/\tau)}}}
} 
\noindent where $N_{\boldsymbol{\tilde{y}}_i}$ is the total number of images in the minibatch that have the same label $\boldsymbol{\tilde{y}}_i$ as the anchor $i$. The total loss is the sum over the loss of all anchors $i$, i.e., $\mathcal{L}=\sum_i\mathcal{L}_i^{sup}$. The loss has important properties well suited for supervised learning: 
\begin{itemize}
    \item generalization to an arbitrary number of positives
    \item contrastive power increases with more negatives.
\end{itemize}

\smallskip

\subsection{Regularization}
Regularization techniques in machine learning (surveyed in \cite{mor20}) have proven very helpful for deep learning. Explicit regularization adds a loss term $R(f)$ for a network $f$ to the loss function $L(x)$ for data $(x_i,y_i)$ with a trade-off parameter $\lambda$.
\eq{\min_f \sum_i L(x_i,y_i)+\lambda R(f)}
Implicit regularization is all other regularization, e.g., early stopping or using a robust loss function. 
Classical $L2$-regularization and dropout\cite{sri14}, where activations of a random set of neurons are set to 0, are among the most wildly used regularization.

\smallskip
\noindent\textbf{$R_1$ Regularization} \label{sec:R1 Re}
\cite{regr1} is used to penalize the  discriminator in generative adversarial networks based on the gradient with the goal of stabilizing training:
\eq{
R_{1}(\psi) = \frac{\gamma}{2}E_{p_{D}(x)}[||\nabla{D_{\psi}(x)}||^{2}]}
Technically, the regularization term penalizes gradients orthogonal to the data manifold.

\smallskip
\noindent\textbf{Entropy Regularization} \label{sec:Entro}\cite{Entro16} aims at fostering diversity. Specifically, asynchronous methods for deep reinforcement learning \cite{wil91,Entro16}. \cite{Entro16} ensures diversity of actions in reinforcment learning, i.e., it prevents overoptimizion towards a small fraction of the environment. The entropy is simply computed over the probability distribution of actions given by the policy $\pi(x)$ as:
 \eq{H(x)=\sum_x \pi(x)\cdot\log(\pi(x))}



\smallskip
\noindent\textbf{Path Length Regularization} \cite{Path19} \label{sec:Path }
for generative adversarial networks aims at ensuring that the fixed-size step length in the latent space matches the fixed-magnitude change in the image. The idea is to encourage that a fixed-size step in the latent space $\mathcal{W}$ results in a non-zero, fixed-magnitude change in the image. The goal is to ensure better conditioning of GANs, simplifying architecture search and generator inversion. Gradients with respect to $\mathbf{w} \in \mathcal{W}$ stemming from random directions in the image space should be almost equal in length independent of $\mathbf{w}$ or the image space direction. The local metric scaling characteristics of the generator $g: \mathcal{W} \rightarrow \mathcal{Y}$ are captured by the Jacobian matrix $\mathbf{J_{w}} = \delta{g}(\mathbf{w})/\delta{\mathbf{w}}$. The regularizer becomes:

\eq{ \mathbb{E}_{\mathbf{w},\mathbf{y} \sim \mathcal{N}(0, \mathbf{I})} (||\mathbf{J}^{\mathbf{T}}_{\mathbf{w}}\mathbf{y}||_{2} - a)^{2}
}
\noindent where $y$ are random images with normally distributed pixel values, and $w \sim f(z)$, where $z$ is normally distributed. The constant $a$ is the exponential moving average of $||\mathbf{J}^{\mathbf{T}}_{\mathbf{w}}\mathbf{y}||_{2}$. The paper further avoids the computationally expensive, explicit computation of the Jacobian.

\smallskip
\noindent\textbf{DropBlock}\cite{DropB18} \label{sec:DropB}
 drops correlated areas of features maps rather than selecting features to drop independently. This is especially suitable for convolutional neural networks where features maps exhibit spatial correlation and a (real-world) feature often corresponds to a contiguous spatial area in feature maps.

\smallskip

\subsection{Optimization} 
Optimization(surveyed in \cite{sun20}) is the process of estimating all network parameters so that the loss function is minimized. The two most wildly known technique is stochastic gradient descent(SGD) and Adam. None strictly outperforms in all cases in terms of generalization performance. SGD dates back at least to the 50ies\cite{kief52}, while Adam stems from 2014\cite{Adam14}.

\smallskip
\noindent\textbf{Adafactor} \label{sec:Adafa}
 \cite{Adafa18} reduces the memory needs of the Adam optimization by maintaining only row- and column-wise statistics of parameter matrixes rather than per-element information.

\smallskip
\noindent\textbf{Layerwise adaptive large batch optimization (LAMB)}\cite{LAMB19} \label{sec:LAMB}
 builds on Adam and accelerates training using large mini-batches. It performs per-dimension and layerwise normalization.



\smallskip
\noindent\textbf{Two Time-scale Update Rule(TTUR)}: \label{sec:TTUR}
For generative adversarial networks trained with stochastic gradient descent TTUR\cite{TTUR17} uses a separate learning rate for the discriminator and generator. For a fixed generator, the discriminator reaches a local minimum. This still holds if the generator converges slowly, e.g., using a small(er) learning rate. This helps in convergence of the GAN and it can improve performance since the generator captures the feedback of the discriminator more profoundly before pushing it into new regions.

\smallskip
\noindent\textbf{Decoupled Weight Decay Regularization for ADAM}: \label{sec:AdamW}
AdamW\cite{AdamW17} is built on a simple observation and implemenation. The orginal Adam optimization changes weights due to (L2-)regularization after computation of gradients for Adam. But intuitively moving averages of gradients should not include regularization.



\smallskip
\noindent\textbf{RAdam and AMSGrad}: \label{sec:RAdam}
Both techniques tackle the convergence problem of Adam. Rectified Adam\cite{RAdam19} rectifies the variance of the adaptive learning rate, which is large initially. Thus, similar to the warm-up heuristic small initial learning rates can help. 
AMSGrad \cite{AMSGr19} uses the maximum of past squared gradients rather than the exponential average.

\smallskip
\noindent\textbf{Stochastic Weight Averaging}: \label{sec:Stoch}
Simple averaging of weights from different epochs during stochastic gradient descent with constant or cycling learning rate improves performance.\cite{Stoch18} 







\smallskip
\noindent\textbf{Sharpness-Aware Minimization}\cite{Sharp20} \label{sec:Sharp}
 minimizes loss value and sharpness, which improves generalization. It finds parameters with neighborhoods of low loss value (rather than parameters that only themselves have low loss value). The loss is:
\eq{
\min_w \max_{||\epsilon||_p\leq \rho} L(w+\epsilon)
}

\smallskip

\section{Self, Semi-supervised and Contrastive learning}
Semi-supervised learning leverages a large amount of unlabelled data based on a small amount of labeled data (see \cite{yang22} for a survey). 
Self-supervised learning benefits from self-generated (pseudo)labels stemming from artificial tasks. Both reduce the burden of collecting (human) labeled data. Self-supervised (pre-)training combined with fine-tuning on a (small) human-annotated dataset can lead to state-of-the-art results. The paradigm has grown extensively in recent years (surveyed in \cite{eric22}). It is commonly combined with contrastive learning. In contrastive learning, the goal is to learn to distinguish between similar and dissimilar data. Since data can be automatically distorted to different extents, creating ``pseudo-labeled'' data for self-supervised learning can be straightforward. 





\smallskip
\noindent The \textbf{simple framework for contrastive learning} (SimCLR)\cite{SimCL20} \label{sec:SimCL}
 maximizes agreement between two inputs that result from augmenting the same data sample differently. Augmentation can be random cropping, color distortions, and Gaussian blur. To obtain reprsentation vectors, a standard ResNet\cite{he16} is used. Representations are further processed using a simple MLP before the contrastive loss is applied.

\smallskip
\noindent\textbf{Bootstrap Your Own Latent (BYOL)} \label{sec:BYOL}
 \cite{BYOL20} uses an online and a target network. Both have the same architecture consisting of an encoder, a projector, and a predictor but they do not share weights. The target network's parameters are an exponential moving average of the online network's parameters. The online network has to predict the target network's representation given an augmentation of the (same) input.


\smallskip
\noindent\textbf{Barlow Twins}\cite{barl19} \label{sec:Barlo}
rely on an objective function that aims to reduce cross-correlation $C$ between outputs for a set of image $Y^A$ and their distorted versions $Y^B$ as close to the identity as possible, i.e., the loss (including  $\lambda$ as a tuning parameter) is:
\eq{L=\sum_i (1-C_{i,i})^2+\lambda\cdot\sum_i\sum_{j\neq i} C_{i,j}^2}

\smallskip
\noindent\textbf{Momentum Contrast (MoCo)} \label{sec:MoCo}
\cite{MoCo19} builds a dynamic dictionary represented by an encoder using unsupervised contrastive learning. Training performs look-ups and enforces that an encoded query should be similar to its matching encoded key and dissimilar to others.
The dictionary is a queue of data samples. For every mini-batch, encoded samples are added, and the oldest mini-batch are dequeud.  The key encoder is a momentum-based moving average of the query encoder, which should help to maintain consistency.


\smallskip
\noindent\textbf{Noisy Student}: \label{sec:Noisy}
The paper\cite{Noisy19} describes training an (EfficientNet) model on labeled data. This model is used as a teacher to generate pseudo labels for unlabeled images. A larger (EfficientNet) model is trained on the union of all data. This process is repeated, i.e., the student becomes the teacher of a new student. During student training, noise such as dropout and data augmentation are applied so that the student's learning is harder and it can improve on the teacher.

\smallskip
\noindent\textbf{FixMatch} \cite{FixMa20} \label{sec:FixMa}
predicts the label of a weakly-augmented image. If the confidence for a label is above a threshold, then the model is trained to produce the same label for the strongly-augmented version of the image.


\smallskip

\section{Architectures and Layers}
We elaborate on four important layers types, i.e., activation-, skip-, normalization-, and attention layers followed by numerous contemporary architectures based on transformers as well as graph neural networks.

\smallskip

\subsection{Activation}
Activation functions  are usually non-linear. They have a profound impact on gradient flow and, thus, on learning. Early activation functions commonly used from the 1960s throuhout the early 2000s such as sigmoid and tanh make training deep networks difficult due to the vanishing gradient when these functions saturate. The introduction of the rectified linear unit $ReLU$ in 2010\cite{nai10} marked a breakthrough result. While its original version is still commonly used, transformer architectures have popularized other activation functions and ReLU variants. Most of them still share qualitatively the behavior of ReLU, i.e., for negative inputs, outputs are of small magnitude and for positive inputs, they are unbounded (see \cite{api21} for a survey). 

\smallskip
\noindent\textbf{Gaussian Error Linear Units (GELU)}\cite{GELU16}  \label{sec:GELU}
weigh inputs by their precentile (ReLUs only use the sign). Activation is the product of the input and the standard Gaussian cumulative distribution function $\Phi(x)$, i.e.,
\eq{GELU(x)= x\cdot \Phi(x)}


\smallskip
\noindent The \textbf{Mish} activation\cite{Mish19} \label{sec:Mish}
originates from systematic experimentation inspired by Swish and ReLU: 
\eq{&f(x)=x\cdot \tanh(soft^+(x))\\
&\text{ with } soft^+(x):=\ln(1+e^x)
}
In comparison, the Swish activation\cite{Swish17}  is:
\eq{&f(x)=x\cdot sigmoid(\beta x) \label{eq:swi}} Here $\beta$ is a learnable parameter.



\smallskip

\subsection{Skip connections}
Skip connections originate from residual networks\cite{he16}. In the simplest form, the output $y$ for an input $x$ of a single layer $L$ (or a set of a few layers) with a skip connection is $y(x)=L(x)+x$.
The original paper used the term residual since the layer $L$ has to learn a residual $L(x)=H(x)-x$ rather than the desired mapping $H$ itself. Since then, skip connections have been used in many variations.

\smallskip
\noindent\textbf{Inverted Residual Block}\cite{Inver18}: \label{sec:Inver}
By inverting the channel width to a narrow-wide-narrow layer sequence from the original wide-narrow-wide order\cite{he16} in combination with depthwise convolutions for the wide-layer, parameters are reduced, and residual blocks execute faster. 

\smallskip
\noindent A \textbf{Dense Block}\cite{Dense16} \label{sec:Dense}
 receives inputs from all prior layers (with matching feature-map sizes) and connects to all subsequent layers (with matching feature-map sizes).

\smallskip

\noindent\textbf{ResNeXt Block}\cite{ResNe16}: \label{sec:ResNe}
This split-transform-merge approach for residual blocks entails evaluating multiple residual blocks in parallel and aggregating them back into a single output.

\smallskip

\subsection{Normalization}
Since the introduction of batch-normalization\cite{iof15}, normalization has been a very successful concept in improving training speed, stability, and generalization of neural networks. However, their need is debated\cite{shao20}, e.g., for some applications careful initialization and adjustments of learning rates might make them at least partially redundant. The idea of normalization is to transform a value $x$ to a normalized value $\tilde{x}$, by subtracting the mean $\mu$ and scaling by the standard deviation $\sigma$, i.e.,
$\tilde{x} = \frac{x-\mu}{\sigma}$.
Normalization approaches differ in the computation of $\mu$ and $\sigma$, e.g., $\mu$ and $\sigma$ can be computed across different channels. 

\smallskip
\noindent\textbf{Layer Normalization}: 
Given summed inputs, normalization statistics are computed\cite{Layer16} for a layer $L$ with $|L|$ neurons as:
\eqs{&\mu = \frac{1}{|L|} \sum_{i=0}^{|L|-1} a_i \text{\phantom{abcd}} \sigma =  \sqrt{\frac{1}{|L|} \sum_{i=0}^{|L|-1} (a_i-\mu)^2} }
In contrast to  batch-normalization, it poses no restrictions on batch size and also no dependencies between batches. In particular, it can be used with batch size 1. 

\smallskip
\noindent\textbf{Instance Normalization}\cite{Insta16} \label{sec:Insta}
computes for a 4-dimensional input, such as an image with height $H$, width $W$, channels $C$, and batch size $T$:
\eq{&\mu_{t,c} = \frac{1}{HWT} \sum_{t<T,w<W,h<H} x_{t,c,w,h}\\
&\sigma_{t,c} =  \sqrt{\frac{1}{HWT} \sum_{t<T,w<W,h<H} (x_{t,c,w,h}-\mu_{t,c})^2}
}
It can be used, e.g., to normalize contrast for an image. There exist multiple versions of it, e.g., a version that scales based on weight norms\cite{Weigh19}. 

\smallskip
\noindent\textbf{LayerScale}\cite{Layer21} 
has been introduced in the context of transformers as a per-channel multiplication of outputs of a residual block with a diagonal matrix:
\eq{
&x_{l'}= x_{l}+diag(\lambda_1,...,\lambda_{d})\cdot SA(\eta(x))\\
&x_{l+1}= x_{l'}+diag(\lambda_1,...,\lambda_{d})\cdot FFN(\eta(x))
}
$SA$ is the self-attention layer, $FFN$ is the feed forward network, and $\eta$ the layer-normalisation (see Figure \ref{fig:att}).

\smallskip

\subsection{Attention}
Attention mechanisms  (surveyed in \cite{bra21,attvis22}) allow for learning relevance scores for inputs, similar to how cognitive attention works. Some parts of the inputs can be deemed highly important, while others are disregarded as irrelevant. The relevance of a particular input can often be determined by contextual information, such as the relevance of a word in a text document often depends on nearby words. 

\smallskip
\noindent\textbf{Scaled Dot-Product Multi-Head Attention} \cite{Scale17}: \label{sec:Scale}
Using dot products combined with down-scaling has proven very successful in computing attention scores. 
Attention takes a query $Q$, a key $K$ and a value $V$ as inputs and outputs an attention score:
\eq{\text{Att}(Q, K, V) = \text{softmax}\big(\frac{QK^{T}}{\sqrt{d_k}}\big)\cdot V}

Using multiple, independent attention mechanisms in parallel allows attending to various aspects of the input. Formally, in multi-head attention, we learn matrixes \textbf{W}:
\eq{
&\text{MultiHead}(\textbf{Q}, \textbf{K}, \textbf{V}) = [\text{h}_{0},\dots,\text{h}_{n-1}]\textbf{W}_{0}\\
&\text{where } \text{head h}_{i} = \text{Att} (\textbf{Q}\textbf{W}_{i}^{Q}, \textbf{K}\textbf{W}_{i}^{K}, \textbf{V}\textbf{W}_{i}^{V} ) }

\smallskip
\noindent\textbf{Factorized (Self-)Attention} \cite{Fixed19} \label{sec:Fixed}
reduces the computational and memory footprint of attention. While (full) self-attention\cite{Scale17} allows attending to every prior input element, factorized self-attention allows only to attend to a subset thereof.
Formally, an output matrix is computed given a matrix of input embeddings $X$ and the connectivity pattern $S=\{S_1,...,S_{n}\}$, where $S_i$ is the set of indices of input vectors attended to by the $i$th output vector.
\eq{
&\text{FacAtt}(X, S) = (A(\mathbf{x}_{i},S_{i}))_{i\in[1,n]}\\
&a(\mathbf{x}_{i}, S_{i}) = \text{softmax}(\frac{(W_{q}\mathbf{x}_{i})K^{T}_{S_{i}}}{\sqrt{d}})\cdot V_{S_{i}} \\
&K_{Si} = (W_{k}\mathbf{x}_{j})_{j\in{S_{i}}} \text{\phantom{abc}} V_{S_i} = (W_{v}\mathbf{x}_{j})_{j\in{S_{i}}} 
}
For full self-attention $S^F_i:=\{j|j\neq i\}$ (indexes to prior inputs to $i$). In contrast, factorized self-attention has $p$ separate attention heads, where the $m$th head defines a subset $A_i^{(m)} \subset S^F_i$  and lets $S_i=A_i^{(m)}$.
For strided self-attention:
\eq{ &A_{i}^{(1)}=\{t,t+1,...i\} \text{ for } t=\max(0,i-l)\\
&A_i^{(2)}=\{j:(i-j)\mod l=0\}}
This pattern is suitable, when structure aligns with the stride-like images. For data without a periodic structure like text, fixed attention can be preferable:
\eq{
&A_{i}^{(1)}=\{j: \lfloor j/l\rfloor = \lfloor i/l \rfloor \}\\
& A_i^{(2)}=\{j: j \mod l \in \{t,t+1,...l\}\}
}where $t=l-c$ and $c$ is a hyperparameter. For example, for stride 128 and $c=8$, all future positions greater than 128 can attend to positions 120-128, all greater 256 to 248-256, etc.

\smallskip
\noindent A \textbf{Residual Attention Network (RAN)}\cite{RAN17} \label{sec:RAN}
 module leverages the idea of skip connections. It consists of a mask and a trunk branch. The trunk branch performs feature processing. It can be any network. The mask branch represents feature weights.
The output of an attention module is
\eq{H_{i,c}(x)=(1+M_{i,c}(x))\cdot F_{i,c}(X) }
Here $i$ is a spatial position and $c$ is a channel. $M(x)$ should be approximatedly 0, $H(x)$ approximates original features $F(x)$.

\smallskip
\noindent\textbf{Large Kernel Attention}\cite{guo22} \label{sec:Visua}
 decomposes a large scale convolution into three smaller scale convolutions using common ideas, i.e., depth-wise dilated convolution, a non-dilated depthwise convolution, and a channel-wise 1x1 convolution. For the output of these convolutions, an attention map is learned.

\smallskip
\noindent\textbf{Sliding Window Attention}\cite{Slidi20} \label{sec:Slidi}
 aims at improving the time and memory complexity of attention. It reduces the number of considered input pairs. More precisely, for a given window size $w$ each token attends to $\frac{w}{2}$ tokens on each side.

\smallskip

\subsection{Transformers } 
Transformers have quickly become the dominant architecture in deep learning. Combined with self-supervised training on large datasets, they have reached state-of-the-art on many benchmarks in NLP(see \cite{liu23} for a survey) and computer vision (surveyed in \cite{han2022,kha22}). Since their introduction in 2017\cite{Scale17} countless versions have emerged that tackle issues of the original transformer such as computational overhead and data efficiency. 

\begin{figure*}
  \centering
  \includegraphics[width=0.5\linewidth]{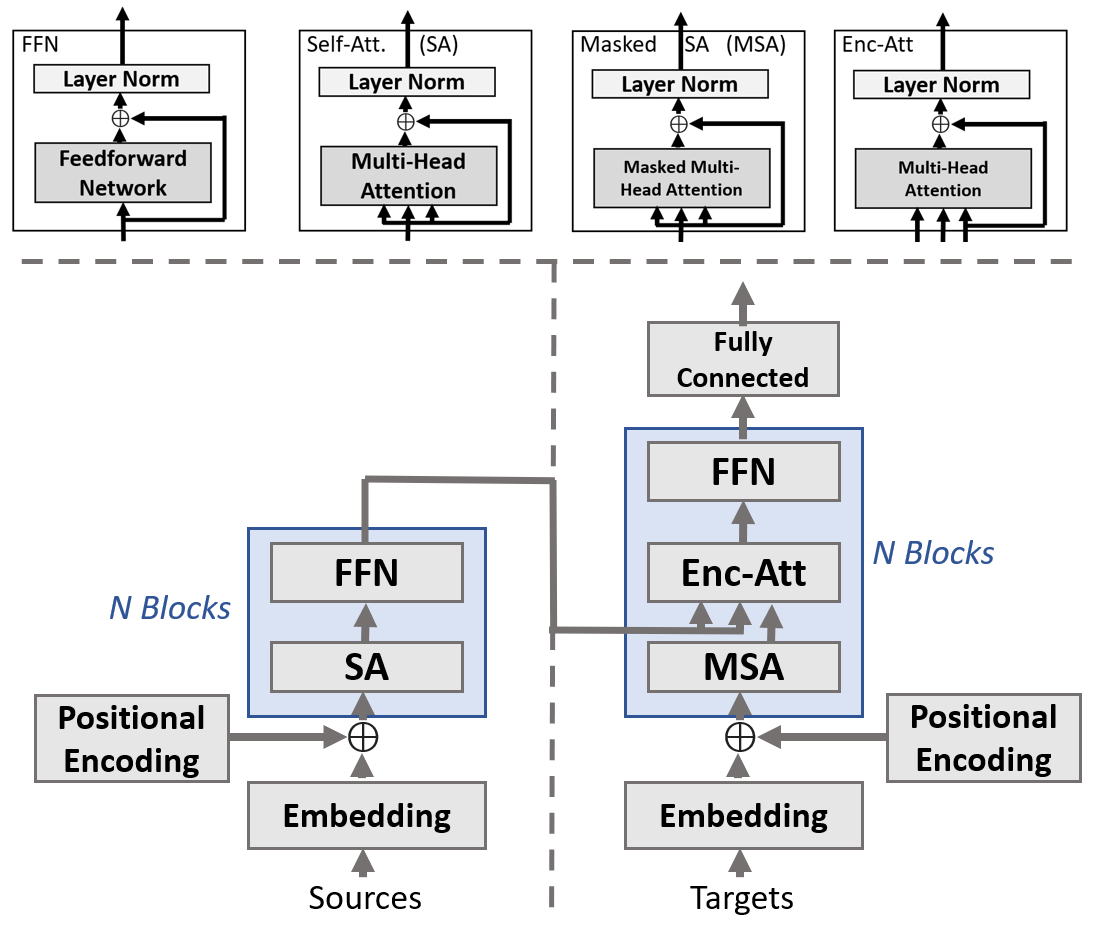}
\caption{Transformer with the four basic blocks on top and the encoder and decoder at the bottom} \label{fig:att}
\end{figure*}

Transformers are said to have less inductive bias and are in turn more flexible than other architectures, such as convolutional neural networks and recurrent networks. Thus, they also require more training data to compensate for the lack of inductive bias. Since large amounts of labeled data are difficult to obtain, transformers are commonly trained using self-supervised learning, i.e., pseudo-labels.
The original transformer\cite{Scale17}, developed for natural language processing, employs an encoder and decoder like earlier recurrent neural networks. It stacks multiple transformer blocks on top of each other, as illustrated in Figure \ref{fig:att}. Key elements are multi-head attention, layer normalization, and skip connections. Furthermore, positional encodings and embeddings of inputs play an important role. The absolute positional encodings $PE$ for position $pos$ in \cite{Scale17} uses sine and cosine functions varying in frequency:
\eq{
&\text{PE}(pos, 2i) = \sin(pos/10000^{2i/d}) \\
&\text{PE}(pos, 2i+1) =  \cos(pos/10000^{(2i)/d}) 
}
where $i$ is the dimension of the encoding and $d$ is the number of dimensions. The choice was motivated by the fact that relative positions, which might be equally relevant to absolute ones, are a linear function of absolute position encodings.

\smallskip
\noindent\textbf{Bidirectional Encoder Representations from Transformers(BERT)} \cite{BERT18} \label{sec:BERT}
yields contextual word-embeddings using the encoder of the transformer architecture. It relies on a masked language model pre-training objective and self-supervised learning. The model must predict randomly chosen, masked input tokens given its context. Thus, the model has bidirectional information, i.e., it is fed tokens before and after the masked words. In classical next-word prediction no tokens after the word to predict are given. As a second prediction task, the model must predict if a sentence pair $(A,B)$ consists of two consecutive sentences $A$ and $B$ within some document (or two possibly unrelated sentences). The pre-trained model based on self-supervised training can be fine-tuned for downstream tasks using labeled data.

The original BERT model has since then improved in many ways, e.g., \cite{san19} reduced the computational burden of BERT, and \cite{RoBER19} trained models longer, on longer sequences, with bigger batches over more data, etc. This led to more robust and generalizable representations.


\smallskip
\noindent\textbf{GPT to GPT-3 on to ChatGPT and GPT-4}: \label{sec:GPT-2}
GPT is based on the decoder of a transformer to predict tokens sequentially. GPT\cite{Impr18} first performs pre-training in an unsupervised way before applying supervised fine-tuning. Pre-training takes place on a large corpus of tokens $U=(u_0,u_1,...,u_{n-1})$ by maximizing the likelihood of the next token given prior tokens:
\eq{L(U)=\sum_i p(u_i|u_{i-k},...,u_{i-1})}where $k$ is the size of the context window and the conditional probability is modeled using a neural network, i.e., using a multi-layer transformer decoder\cite{liu18gen} by dropping the encoder in \cite{Scale17}. Rather than only predicting the next token given an input, the model is also trained to predict input tokens. Furthermore, the memory footprint of attention is lowered.
GPT-2 \cite{GPT-219} builds on GPT with few modifications, e.g., layer normalization locations were changed (moved to the input of each sub-block, and an extra normalization was added after the final self-attention block), initialization of residual weights was scaled, and the vocabulary, context, and batch size were increased.
GPT-3's\cite{GPT-320} architecture is almost identical to that of GPT-2, but the number of parameters is more than 100 times larger and it differs in (amount of) training data. ChatGPT\cite{Chat22} is a sibling to InstructGPT\cite{ouy22}, which is optimized towards following user intentions. InstructGPT applies fine-tuning of GPT-3 in a two-step process: (i) based on labeler demonstrations through supervised learning and (ii) based on human rankings of model outputs using reinforcement learning.
ChatGPT  follows the same procedure, i.e., (i) for supervised learning, human AI trainers provided conversations by playing both the human user and the AI assistant. The resulting dialogue dataset was enhanced with the InstructGPT dataset, which was transformed into a dialogue format. (ii) Conversations of AI trainers with ChatGPT were ranked, i.e., for a randomly selected model-written message, AI trainers ranked several alternative completions. The ranking dataset was used for reinforcement learning. 
The process was repeated multiple times. \\
Technical details of the successor of ChatGPT, i.e., GPT-4 have not been disclosed\cite{gpt423}. The provided technical report indicates that it is similar to ChatGPT. GPT-4 is multi-modal, i.e., it can also process images, however, details are unknown. The report only points towards major improvements in training efficiency. The accomplishment was to predict the performance of large scale models using the performance of small models (possibly trained on less data). This is highly important as computational costs and time can be a key factor for large deep learning models.

\smallskip
\noindent\textbf{Text-to-Text Transfer Transformer} (T5)\cite{T520} \label{sec:T5}
 views every text-based language models as generating an output text from a given input text. It differs from BERT\cite{BERT18} by using causal masking during training for predicting the target. Causal masking prevents the network from accessing ``future'' tokens of the target. T5 also differs in pre-training tasks.

\noindent\textbf{BART}\cite{BART19} is a denoising autoencoder for pretraining sequence-to-sequence models that uses a standard transformer based machine translation architecture. It has been shown to be effective for language generation, translation, and comprehension. Training is based on corrupting text with noising functions ranging from token deletion, masking onto sentence permutation and document rotation. Learning stems form reconstructing the original text from its corrputed version. The flexibility in noising options is attributed due to BART's generalization of prior works such as BERT and GPT, i.e., the encoder is bi-directional (like BERT), while the decoder is autoregressive (like GPT). 

\smallskip

\noindent\textbf{XLNet} \label{sec:XLNet}
\cite{XLNet19} combines advantages of autoregressive modeling like GPT, predicting the next token, and denoising auto-encoding BERT\cite{BERT18}, reconstructing $x$ given a noisy input $\hat{x}$ that originates through masking words of $x$. It does so by using a permutation language model that samples a permutation of $Z={z_0,z_1,...,z_{T-1}}$ of the sequence $(0,1,2,...,T-1)$ leading to the objective:
\eq{\max p(u_{z_T}|u_{z_0},...,u_{z_{T-1})}}
There is no actual permutation of inputs, which would be unnatural (and not occurring during later fine-tuning tasks). Rather, the permutation impacts the attention mask to ensure that the factorization order by $Z$ is maintained. 

\smallskip
\noindent The \textbf{Vision Transformer} \cite{doso20} \label{sec:Visio}
 relies heavily on the original transformer. An image is partitioned into small patches, which are flattened and linearly embedded with position embeddings. A standard transformer encoder then processes the created vector of each patch.

\smallskip
\noindent The \textbf{Swin Transformer} \cite{Swin21} \label{sec:Swin }
 for computer vision builds hierarchical feature maps rather than just a single (resolution) feature map. It also only computes self-attention within a local window reducing computation time.




\noindent\textbf{PaLM (2):} The original PaLM\cite{PaLM22} is a large language model consisting of 540 billion parameters similar to other more prominent such as GPT-3. Technical innovation discussed is mostly on the scaling of model training, i.e., a single model can be trained across  tens of thousands of accelerator chips efficiently. The original transformer architecture\cite{Scale17} is also adjusted slightly, e.g., SwiGLU activations are used, i.e., \eq{Swish(xW)\cdot xV }, where Swish is given by Eq. \ref{eq:swi}, different positional embeddings (better for long sequences), and multi-query attention (faster computation), no biases (better training stability), and shared input-output embeddings. \\ 

PaLM 2\cite{PaLM23} is the better performing successor of PaLM that differs in terms of dataset mixtures, e.g., using more diverse languages as well as domains (e.g., programing languages, mathematics). It also uses the classical transformer architecture. However, it uses a smaller model than the first PaLM version but more training compute. It also relies on more diverse pre-training objectives (than simple next word or masked word prediction).

\smallskip

\subsection{Graph Neural Networks }
Graph neural networks (surveyed in \cite{wu2020}) can be seen as a generalization of CNNs and transformers. They operate on graph data, i.e., nodes connected with edges. We discuss graph models, including models to obtain node embeddings that can be used for downstream tasks.

\smallskip
\noindent\textbf{Graph Convolutional Networks}\label{sec:GCN}
\cite{GCN16} use CNNs for semi-supervised learning. They approximate spectral graph convolutions using polynomials of order $k$, which a CNN can compute with $k$ linear layers. 

\smallskip
\noindent\textbf{Graph Attention Networks}\label{sec:GAT}
\cite{GAT17} rely on masked self-attention layers allowing nodes to attend flexibly over their neighborhoods' features, i.e., node $j$ obtains importance scores for node $i$'s features. Masking allows to only consider edges between node pairs that are actually connected. In contrast to GCN, different importances for nodes in the same neighborhood can be assigned. Also, it does not rely on costly matrix operations for eigendecompositions.

\smallskip
\noindent\textbf{Graph Transformer}\label{sec:Graph}
\cite{Graph20} extends the original transformer to graphs by using attention over neighborhood connectivity for each node, generalizing the position encoding, replacing layer- with batch-normalization, and learning edge representations (in addition to node representations).

\smallskip
\noindent\textbf{TuckER}\label{sec:TuckE}
\cite{TuckE19}  performs factorization for link prediction in knowledge graph. Knowledge is represented as (subject, relation, object) triplets, and the task is to predict whether two entities are related.  The graph can be represented as a binary tensor with the subjects, relations, and objects as dimensions. They use Tucker decompositions to decompose the binary tensor into a product of a core matrix and embedding matrices for subjects, relations, and objects.

\smallskip
\noindent\textbf{Embedding by Relational Rotation} \label{sec:Rotat}
 (RotatE)\cite{Rotat19} performs missing link prediction in knowledge graphs (like the priorly described TuckER\cite{TuckE19}) to model more relational properties such as composition and inversion. They embed entities into a complex space and treat the relation as an element-wise rotation that is optimized to lead from one entity to the other.

\smallskip
\noindent\textbf{Scalable Feature Learning for Networks}\label{sec:node2}
 (Node2Vec)\cite{node216}  learns feature vectors that preserve a node's neighborhood. They use random walks to generate sample neighborhoods, thereby, nodes are viewed based on their role or communities they belong to.

\smallskip

\section{Discussion}
Our survey focused on key design elements in building deep learning models. Taking a practical approach, we chose to ignore theoretical works, which should be further explored in future studies. 
Our findings suggest that despite many small and creative innovations since the original transformer architecture, there have not been any significant "breakthrough" discoveries that have led to much better leaderboard results. The last few years have been characterized by the enlargement of existing networks such as GPT, the increase of data volume (and quality), and a shift towards self-supervised learning.  This could indicate a need for more daring approaches to research rather than incremental improvements of existing works. Combining different elements as outlined in this work could be one way to achieve this.\\

In addition, we noted a few general patterns that have been proven effective in many areas: 
\begin{itemize}
    \item ``Multi-X'', i.e., using the same element multiple times in parallel, such as using multiple residual blocks (ResNeXt) or multi-head attention. This idea is also closely related to ``ensemble learning''.
    \item ``Higher order layers'', i.e., classical CNNs and MLPs only apply linear layers and simple ReLU, but layers like Mish or attention layers perform more complex operations.    
    \item ``Moving average'', i.e., averaging weights such as for SGD and BYOL.
    \item ``Decompose'', i.e., decomposing matrixes such as for TuckER and large kernel attention. 
    \item ``Weighing functions'', i.e., using parameterized weighing functions of inputs can be seen within the attention mechanism but also for GELU units. Therefore, rather than naively aggregating inputs, inputs are weighed and aggregated. The weight might stem from a function with learnt parameters. Such functions can also be seen as ``gates'' that only permit the flow of information within some range of the input parameters.
\end{itemize}
Our survey was also deliberately geared towards more recent works, but still well-established works; this could be perceived as a strength or as a limitation. 
The selection of papers and areas was driven by a prominent platform providing leaderboards. While a reader looking for ``what works well and what is very promising'' benefits from this approach, it could potentially leave out works with exciting ideas that require more research to reveal their full capabilities.  This could be seen as perpetuating the "winner-takes-all" paradigm that reinforces already successful ideas. However, due to the sheer amount of papers, a selection is necessary for conducting a holistic survey of deep learning. 
We acknowledge that online platforms providing leaderboards etc. are very beneficial to the research community and that they should be further advanced. Still, we found that manual verification (e.g., by double checking relevance with Google scholar citations and by reading surveys and papers) was required as we identified works and methods that were not listed correctly on the platform.

\smallskip

\section{Conclusions}
We have presented a brief but comprehensive overview of the deep learning design landscape. We have summarized key works from various significant areas that have emerged in recent years. We believe that our holistic overview in one paper can establish connections that could inspire novel ideas. We have also identified four patterns that characterize many improvements. To further advance the development of deep learning, we need to generate fundamentally new and successful approaches, as the improvements made in the past few years were numerous and often very creative but mainly incremental.



 



\bibliographystyle{apalike}
{\small
\bibliography{refs}}



\end{document}